\documentclass[11pt]{article}
\pdfoutput=1

\usepackage[final]{acl}

\usepackage{times}
\usepackage{latexsym}
\usepackage{adjustbox}
\usepackage[T1]{fontenc}

\usepackage[utf8]{inputenc}
\usepackage{graphicx}
\usepackage{booktabs}
\usepackage{microtype}

\usepackage{inconsolata}
\usepackage{svg}

\title{Towards Optimizing and Evaluating a Retrieval Augmented QA Chatbot using LLMs with Human-in-the-Loop}

\author{Anum Afzal, Alexander Kowsik, Rajna Fani, Florian Matthes \\
School of Computation, Information and Technology \\
  Technical University of Munich \\
  \texttt{ \{anum.afzal, alexander.kowsik, rajna.fani, matthes\}@tum.de} \\}

\begin{document}
\maketitle
\begin{abstract}

Large Language Models have found application in various mundane and repetitive tasks including  Human Resource (HR) support. We worked with the domain experts of SAP SE to develop an HR support chatbot as an efficient and effective tool for addressing employee inquiries. We inserted a human-in-the-loop in various parts of the development cycles such as dataset collection, prompt optimization, and evaluation of generated output. By enhancing the LLM-driven chatbot's response quality and exploring alternative retrieval methods, we have created an efficient, scalable, and flexible tool for HR professionals to address employee inquiries effectively. Our experiments and evaluation conclude that GPT-4 outperforms other models and can overcome inconsistencies in data through internal reasoning capabilities. Additionally, through expert analysis, we infer that reference-free evaluation metrics such as G-Eval and Prometheus demonstrate reliability closely aligned with that of human evaluation.
\end{abstract}

\section{Introduction}

In recent years, incorporating Artificial Intelligence (AI) into various sectors has led to significant improvements in automated systems, particularly in customer service and support. Since the offset of Large Language Models (LLMs), more companies are now incorporating Natural Language Processing (NLP) techniques to minimize the need for human support personnel, especially domain experts \cite{rag}. With a chatbot providing accurate and comprehensive responses promptly, domain experts can redirect their focus towards higher-value tasks, leading to potential cost savings and improved productivity within the HR department. Moreover, an effective chatbot can play a pivotal role in enhancing overall employee satisfaction and engagement by delivering timely and relevant assistance.

To this end, we worked with a SAP SE on developing an HR chatbot to evaluate the potential of LLMs on industrial data. We used domain experts as \textit{a human-in-the-loop} through various iterations of LLM-centric development such as dataset collection, prompt optimization, and most importantly the evaluation of model outputs.

The well-known Retrieval Augmented Generation (RAG) \cite{retrieval_augmented} approach is ideal for this use case as it allows the model to produce more grounded answers, hence reducing hallucinations. We optimized different modules of the standard RAG pipeline such as the retriever and model prompts, while constantly incorporating feedback from the domain experts. While the retrieval accuracy of an LLM could still be assessed to a degree, the generative nature of LLMs makes evaluation of the generated output quite challenging. To overcome this, we explored the effectiveness of both traditional reference-based and reference-free (LLM-based) automatic evaluation metrics while using human evaluation as a baseline. 

We benchmark OpenAI's models in our experiments while using the open-source LongT5 \cite{guo-etal-2022-longt5} and BERT \cite{devlin2019bert} as a baseline. In essence, both the industry and the research community could benefit from our findings related to the retriever and the reliability of automatic evaluation metrics.

\section{Corpus}
The dataset used in the development of the HR chatbot was compiled using SAP's internal HR policies with the help of domain experts. While each sample forms a triplet consisting of a Question, Answer, and Context, additional metadata such as the user's region, company, employment status, and applicable company policies were also included. A snippet of such a sample is shown in Appendix \ref{appendix:dataset_example}. 
The dataset was compiled using two separate sources to have a mix of a gold dataset (FAQ dataset) and a user-utterance dataset (UT dataset). Both datasets follow the same structure and differences exist in the distribution of the questions. We extracted all unique HR articles to form a knowledge base for answering new user questions. Additionally, an evaluation set of 6k samples was used to evaluate both the retriever and the chatbot as a whole.

\subsection{Dataset Collection}
\noindent \textbf{FAQ Dataset (N$\approx$48k):}
This is a collection of potential questions, along with their corresponding articles and gold-standard answers. It is carefully created and curated by domain experts based on the company's internal policies.

\noindent \textbf{UT Dataset (N$\approx$41k):}
This is a collection of real user utterances (UT) gathered from previous iterations of the chatbot. Inspired by a semi-supervised learning approach, a simplistic text-matching approach was implemented that mapped each user query to a question from the FAQ dataset. The chatbot logs from this development cycle were inspected and corrected by the domain experts.


\subsection{Dataset Statistics}
\label{dataset_statistics}
\begin{figure}[]
  \centering
  \includegraphics[width=0.52\textwidth]{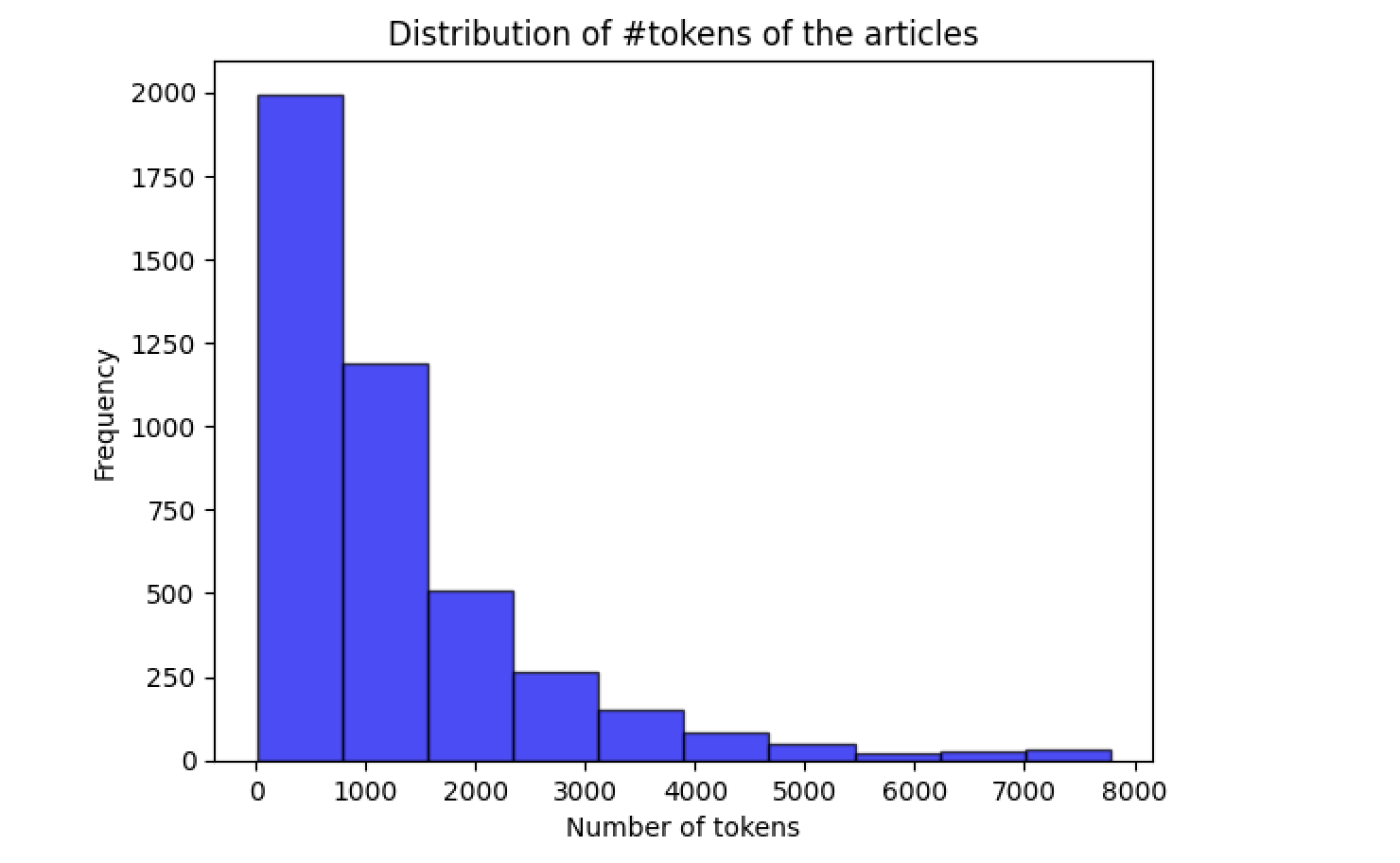}
  \caption{Distribution over the number of tokens of all unique articles in our HR dataset.}
  \label{fig:n_tokens_articles}
\end{figure}

\autoref{fig:n_tokens_articles} shows that the majority of the articles in our dataset have under 4k tokens. Hence, they can easily fit into the context window of OpenAI models. As displayed in \autoref{tab:top10-queries}, the most asked questions in the dataset revolve around payslips, leave days of any kind, and questions regarding management.

\begin{table}[h]
\centering
\resizebox{\columnwidth}{!}
{
\begin{tabular}{p{8cm}l}
\textbf{10 most frequent user queries}\\
\toprule 
How can I change my approver? \\
Where do I see how much leave I have left? \\
How can I view my payslip online?\\
Am I paid during maternity leave?\\
If I am sick whilst on holiday, can I claim my holiday back?\\
Can I cancel a leave request?\\
I have a question about my payslip, who do I contact? \\
Where can I find information about my payslip?\\
Do I receive sick pay?\\
How can I have an overview of my leave?\\

\end{tabular}
}
\caption{Top 10 most frequent user queries}

\label{tab:top10-queries}

\end{table}

\section{Methodology}
\begin{figure*}[h]
    \centering
    \small
    \begin{adjustbox}{width=\textwidth}
        \includegraphics{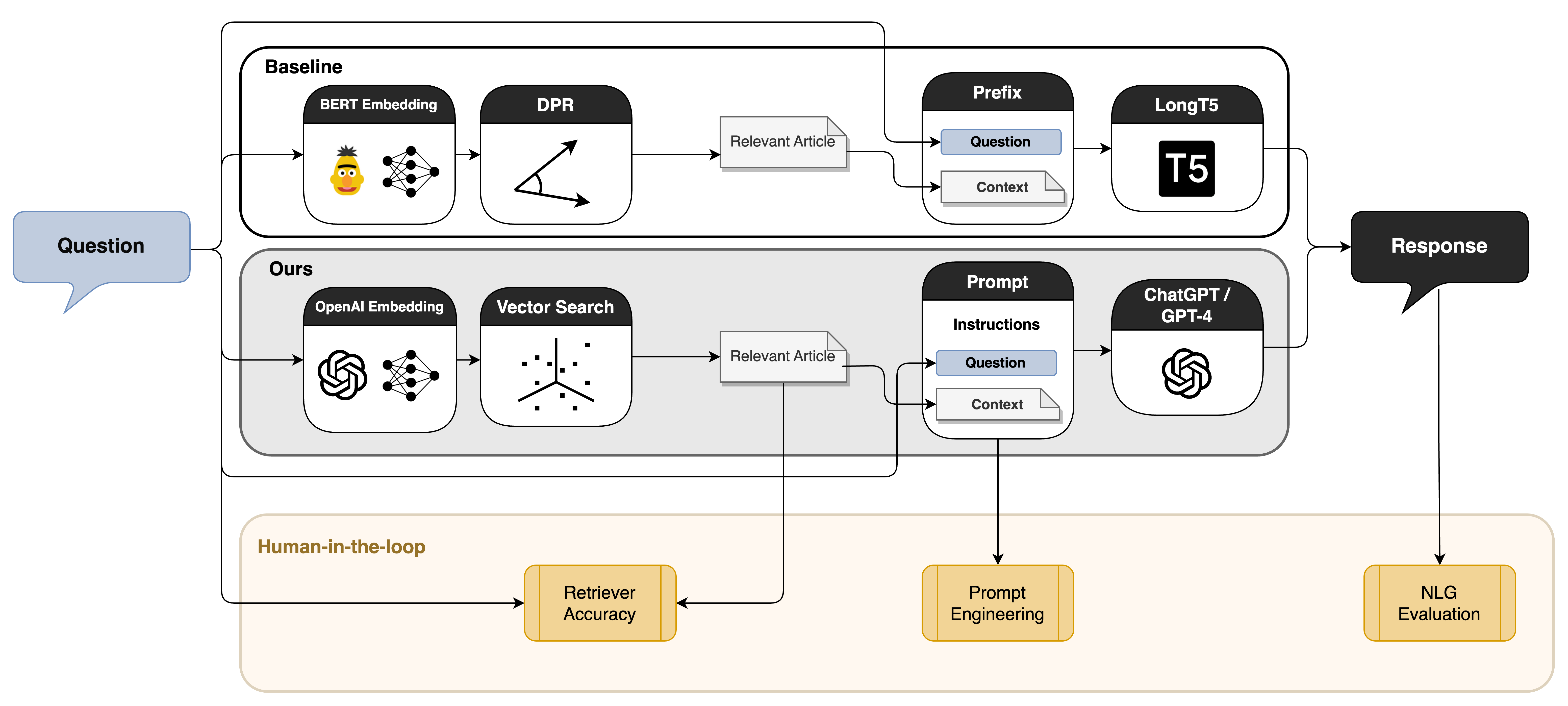} 
    \end{adjustbox}
    \caption{Block diagram of the methodology introduced in our paper, illustrating baseline and Open AI models, highlighting the role of the human-in-the-loop during development}
    \label{fig:methodology}
\end{figure*}

In general, the HR chatbot follows the standard RAG pipeline with optimizations done on individual modules with the help of domain experts as shown in \autoref{fig:methodology}. The methodology illustrates various parts of the chatbot pipeline that are influenced by a human-in-the-loop and is further discussed in Appendix \ref{appendix:human-in-the-loop}.

\subsection{Retriever}
We compiled a comprehensive knowledge base of all possible HR articles occurring in the whole dataset as the basis for retrieval, resulting in roughly 50k unique articles. Given a user utterance, the goal of the retriever is to find the most relevant article from the collection. While the technical details for each retriever may differ, in general, they are both embedding-based. Technical details of the Retriever module are discussed in Appendix \ref{app:dpr-training}.

Moreover, we developed extensive filter functionalities, ensuring that the vector search only considers articles relevant to the user, like their country, region, or employment status as shown in \autoref{tab:dataset_sample}. For example, from the top retrieved articles, we filter them to only keep the ones that are applicable to the employee and then pick the article with the maximum similarity score from the filtered list.

\subsubsection{Dense Passage Retriever (BERT)}
Dense Passage Retriever (DPR) fine-tunes \textit{bert-base-uncased} embedding to generate a model that given a user query, retrieves the most relevant article from a set of documents. The dataset used for training was processed to contain questions paired with their respective gold answers, as well as positive and negative contexts for each question. A triplet loss function \cite{hoffer2018deep} was used for training such that the relevant article served as the positive context, with two random articles from the entire dataset providing the negative contexts. This retriever is used in the framework with the fine-tuned LongT5 model and also serves as a baseline for evaluating the OpenAI retriever.

\subsubsection{Vector Search (OpenAI)}
The OpenAI Retriever is plain vector search, that utilizes the \textit{text-embedding-ada-002} embedding model by OpenAI to generate embeddings for each article, followed by using similarity search to find the relevant article. To further enhance retrieval accuracy, we implemented various \textbf{Query Transformation} techniques\footnote{\url{https://docs.llamaindex.ai/en/stable/optimizing/advanced_retrieval/query_transformations/}} \cite{multi-query}. These methods alter the user query into a different representation using LLMs before the embedding model computes the query vector. The following three query transformation methods were explored and evaluated:

\noindent \textbf{1) Intended Topics:} Inspired by \citet{ma2023query}, the user question is sent to an LLM with the instruction to return a list of three intended topics of the question, which are then embedded instead of the user question. 

\textbf{Example:} \textit{How to request a parental leave?}\\$\rightarrow$ \textit{parental leave, childcare leave, maternity leave}


\noindent \textbf{2) HyDE (Hypothetical Document Embeddings):} In this method introduced by \citet{hyDE}, the user question is transformed by an LLM into three distinct excerpts from potential HR articles answering the original question. These parts are then embedded instead of the user question itself. This approach leads to query embeddings that are very close to the article embeddings, because of the very similar content. 

\textbf{Example:} \textit{How to request a parental leave?} \\$\rightarrow$ \textit{To request parental leave, please submit...}, \textit{If you wish to request...}, ...

\noindent \textbf{3) Multi-Query:} This method\footnote{\url{https://docs.llamaindex.ai/en/latest/examples/retrievers/reciprocal_rerank_fusion/}} employs LLMs to generate multiple variations of a user's question varying in length and phrasing but maintaining the same meaning and intent as the original question. We then embed each of these variants individually. Along with the embedded original question, we perform a vector search for each query, combining the results using Reciprocal Rank Fusion \cite{reciprocal_rank_fusion}. Additionally, we include queries from the \textit{Intended Topics} and \textit{HyDE} methods.


\textbf{Example:} \textit{parental leave request?} \\$\rightarrow$ \textit{How can I request a parental leave?}, \textit{Where can I apply for parental leave?}, ...

\subsection{NLG Module}

\subsubsection{LongT5 (Fine-tuning driven)}
We fine-tuned LongT5 \cite{guo-etal-2022-longt5}, employing the local-attention-based variant\footnote{\url{https://huggingface.co/google/long-t5-local-base}}, which consists of 296 million trainable parameters. This model was fine-tuned on a combination of the FAQ dataset and UT dataset for a generative question-answering task. To limit computational requirements, we fine-tuned it on a context window of 7168 tokens, retaining approximately $\sim$86K samples from the original dataset to avoid truncation. 

\subsubsection{OpenAI Models (Prompt driven)}
We used OpenAI's ChatGPT and GPT-4 to generate the answer to the user's query by passing both the user query and the retrieved article via a meaningful prompt. We conducted extensive prompt engineering to tailor the responses of the LLMs to the company's requirements for an HR chatbot. Prompt engineering was an iterative process that included our qualitative analysis and multiple small evaluations of 10-100 sample responses by the company's HR experts who served as the \textit{human-in-the-loop}. We analyzed feedback from these evaluation runs and addressed the main issues in the next iteration of the process to produce the final prompt shown in \autoref{tab:chatbot_prompt}. 

\subsection{Evaluation Framework}
For our analysis we employ Reference-based evaluation metrics such as BERTScore \cite{bertscore}, ROUGE \cite{rouge}, and BLEU \cite{bleu}. We also explore the concept of using LLM as an evaluator, and finally, we assess the effectiveness of automated metrics by involving domain experts in a human-in-the-loop process. 
\subsubsection{Retriever Evaluation}
Our primary evaluation metric for the retriever is accuracy, defined as the percentage of times the retriever returns the correct article for a given question.
\subsubsection{Human Evaluation Setup}
The domain experts who served as the human-in-the-loop brought a high level of precision and insight to the evaluation process. Apart from dataset curation, they also evaluated the performance of the retriever by verifying the correctness of the retrieved articles. After discussion with domain experts, we found four dimensions across which the quality of the model's output could be evaluated on a score between 1 - 5 following a 5-point Likert \cite{likertScore} scale. One domain expert evaluated 100 samples across the fine-tuned LongT5, ChatGPT and GPT-4 across \textit{Readability}, \textit{Relevance}, \textit{Truthfulness}, and \textit{Usability}.

\subsubsection{Reference-based Metrics}
In evaluating the effectiveness of reference-based metrics, we examine two distinct categories: N-gram-based and embedding-based metrics metrics. 

\noindent\textbf{N-gram based metrics:}
N-gram-based metrics, such as BLEU (Bilingual Evaluation Understudy) and ROUGE (Recall-Oriented Understudy for Gisting Evaluation), assess the similarity between the generated response and the ground truth answer by analyzing the overlap of n-grams.

\noindent\textbf{Embedding-based metrics:}
Embedding-based metrics, such as BERTScore, leverage deep contextual embeddings from language models like BERT to assess the semantic similarity between generated and reference texts. 

\subsubsection{Reference-free Metrics}
In the evolving landscape of Natural Language Generation evaluation, LLM-based metrics emerge as a compelling alternative, offering insights into model performance without the constraints of pre-defined reference responses. Details regarding the prompts used for these Reference-free metrics are present in Appendix \ref{appendix: prompts}.

\noindent\textbf{Prompt-based Evaluation:}
Prompt-based evaluation is at the forefront of NLG advancements, particularly with the utilization of LLMs \cite{leveraging}. Inspired by \textbf{G-Eval}, we followed the approach described by \citet{gpteval} and tailored the prompts to be suitable for the evaluation of a question-answering task. 

\noindent\textbf{Tuning-based Evaluation:}
Nowadays, there is a significant shift toward leveraging open-source language models, such as LLaMA \cite{llama}, for fine-tuning purposes. We utilize \textbf{Prometheus} \cite{prometheus}, which stands out for its fine-tuned evaluation capability, leveraging a large language model to perform nuanced analysis based on customized score rubrics \cite{leveraging}. This unique approach enables Prometheus to evaluate text generation tasks comprehensively, considering factors such as creativity, relevance, and coherence without relying on reference texts.

\section{Results and Discussion}

\subsection{Dense Passage Retriever}
As depicted in \autoref{tab:evaluation_retriever}, surprisingly the BERT-based DPR significantly outperforms all new methods with a top-1 accuracy of 22.24\%, whereas the OpenAI-based retriever only reaches a top-1 accuracy of 11.12\%. Of the latter, the best performer is \textit{Multi-Query}, with 10.92\%, yet this still falls short of the \textit{Basic} retriever (no query transformation). These results resonate with the findings of \citet{weller-etal-2024-generative}, confirming that query transformations, do not always lead to better performance. Our understanding is that the retriever performs poorly mainly because of the noise attributed to the dataset. It is worth noting, that our dataset contains many variant articles for a given topic or question, with only small differences such as the region or the employee role. Hence, the incorrect article may still contain sufficient knowledge to address user queries. We confirmed these findings with our domain experts and elaborated on them further in Appendix \ref{app:dataset-challenges}. Further results on up to top-5 articles are shared in Appendix \ref{sec:topk-evaluation_retriever}.

\noindent However, to assess the effectiveness of the newly implemented methods on a different dataset, we gathered 10k samples from CQADupStack English \cite{CQADupStack}, a collection of English language questions and their top answers from the Stackexchange English forum. We used the same embedding model as the HR dataset to embed this new data and evaluated its top-1 accuracy. It can be observed that the \textit{Intended Topics} method and \textit{HyDE} both underperform compared to the \textit{Basic} retriever. However, the \textit{Multi-Query} method did produce a higher top-1 accuracy. During our experiments, we noticed that these methods are greatly influenced by the choice of query transformation prompts. For instance, when \textit{HyDE} responses closely matched the desired replies, the accuracy was significantly higher. These methods also achieved higher accuracies than the \textit{Basic} on other types of data, which indicates that the performance is also dependent on the type of data used. This might explain why these methods couldn't achieve higher accuracy on the HR dataset.

\begin{table}[]
\centering
\resizebox{\columnwidth}{!}{%
\begin{tabular}{c|c|c}
& \multicolumn{1}{c|}{HR Test Dataset} & \multicolumn{1}{c}{Stackexchange English}\\
\hline
\textbf{Method} & \textbf{top-1}& \textbf{top-1}\\
\hline
BERT-based DPR & \textbf{22.24\%} & -\\
\hline
Basic & \textbf{11.12\%} & 69.5\%\\
Intended Topics & 9.33\% & 57.25\%\\
HyDE & 10.01\% &  65.91\%\\
Multi-Query & 10.92\% & \textbf{71.31\%}\\
\hline
\end{tabular}
}
\caption{Retriever accuracy on the HR test data and the Stackexchange benchmark dataset for various retriever methods on top-1 retrieved articles}
\label{tab:evaluation_retriever}
\end{table}

\subsection{NLG Evaluation}
We use the previously optimized DPRs with the top-1 article for our NLG Module consisting of ChatGPT, GPT-4 and fine-tuned LongT5 as shown in \autoref{fig:methodology}. An overview of all evaluation scores highlighting model performance across several dimensions is summarized in \autoref{tab:benchmark}.  
  
\begin{table}[ht]
\resizebox{\columnwidth}{!}{%
\centering

\footnotesize
\begin{tabular}{lccc}

\textbf{Metric} & \textbf{ChatGPT} & \textbf{GPT-4} & \textbf{LongT5} \\ \hline
\multicolumn{4}{c}{\textit{Reference-based Evaluation}} \\
\midrule
BLEU Score & 0.27 & 0.28 & \textbf{0.41} \\
ROUGE-1 & 0.48 & \textbf{0.52} & 0.51 \\
ROUGE-2 & 0.36 & 0.35 & \textbf{0.43} \\
ROUGE-L & 0.46 & \textbf{0.50} & 0.49 \\
BERTScore\_P & 0.88 & 0.90 & \textbf{0.91} \\
BERTScore\_R & \textbf{0.96} & 0.93 & 0.91 \\
BERTScore\_F1 & 0.90 & \textbf{0.91} & 0.90 \\
\midrule
\multicolumn{4}{c}{\textit{Reference-free Evaluation (LLM-based)}} \\
\midrule
G-Eval: Relevance & 4.03 & \textbf{4.51} & 3.17 \\
G-Eval: Readability & 4.26 & \textbf{4.49} & 3.52 \\
G-Eval: Truthfulness & 4.12 & \textbf{4.80} & 3.36 \\
G-Eval: Usability & 4.67 & \textbf{4.79} & 3.29 \\ 
Prometheus: Relevance & 3.25 & \textbf{3.70} &  2.83 \\
Prometheus: Readability & 3.07 & \textbf{4.22} & 3.73 \\
Prometheus: Truthfulness & 3.20 & \textbf{3.75} & 3.32 \\
Prometheus: Usability & 3.98 & \textbf{4.32} & 2.83 \\
\midrule
\multicolumn{4}{c}{\textit{Domain Expert Evaluation}} \\
\midrule
Human Eval: Readability & 4.31 & \textbf{4.76} & 4.02 \\
Human Eval: Relevance & 4.31 & \textbf{4.67} & 3.46 \\
Human Eval: Truthfulness & 4.09 & \textbf{4.41} & 3.67 \\
Human Eval: Usability & 3.32 & \textbf{4.11} & 2.59 \\
\hline
\end{tabular}
}
\caption{Average Evaluation Scores. BLEU (0 to 1), ROUGE (0 to 1) and BERTScore (-1 to +1 ) were computed on 200 samples, Prometheus (1 to 5) on 60 samples, and Domain Expert Evaluation (1 to 5) \& G-Eval (1 - 5) on 100 samples.}
\label{tab:benchmark}
\end{table}

\noindent Overall, GPT-4 shows clear domination in terms of generation capabilities for an HR chatbot. N-gram-based evaluation scores such as ROUGE and BLEU are quite low due to the generative nature of the (L)LMs, as the answer may contain words different than the reference answers. Nonetheless, these results establish GPT-4 as the leading model, effectively combining advanced language skills with the demands of content accuracy and user engagement. On the other hand, the fine-tuned LongT5's performance is observed to be inferior when benchmarked against the OpenAI models. This outcome is consistent with the anticipated advancements in LLMs, which are progressively outpacing the capabilities of fine-tuning-driven models. The performance of ChatGPT has been notably strong, trailing marginally behind GPT-4 in only a few scoring categories. Its close performance to GPT-4 raises important considerations for the trade-offs between computational efficiency and output quality.

\subsection{Correlation Analysis}
Inspired by \citet{correlation}, we assessed the reliability of the evaluation score using Spearman \cite{spearman} and Kendall \cite{kendall} correlation coefficients in \autoref{tab:metrics-correlation}. 

\noindent\textbf{
Human Evaluation \& Reference-based Metrics} 

\noindent Due to its limited innovation, LongT5 typically produces text with fewer novel sentences, resulting in more favorable scores from n-gram-based metrics like BLEU and ROUGE. The analysis of GPT-3.5 and GPT-4, in particular, illuminates a significant gap between automated metrics and human judgment. As these models generate more varied and longer sentences, their outputs increasingly diverge from the patterns recognized by word-overlap metrics, such as BLEU and ROUGE. For instance, GPT-4’s BLEU score correlation marks a clear disconnect, indicating that as text generation becomes more complex, the less effective traditional metrics are in evaluating it. This discrepancy calls into question the reliance on current automated metrics for assessing the creativity and nuance of outputs from advanced language models, highlighting the need for more sophisticated evaluation frameworks that can better align with human judgment.

\noindent\textbf{Human Evaluation \& Reference-free Metrics}

\noindent Despite similar average scores between Reference-free metrics and Domain Expert evaluations shown in \autoref{tab:benchmark}, their correlations are low. Since these methods measure linear and ordinal relationships, similar averages in evaluations do not imply a strong correlation as depicted in \autoref{tab:metrics-correlation}.

Overall, while Prometheus and G-Eval both serve as proxies for human evaluation, their effectiveness varies by model and evaluation criteria. While G-Eval excels in assessing truthfulness, its capability in evaluating readability and usability lags behind. Prometheus on the other hand, outperforms G-Eval in assessing usability across all models. However, G-Eval shows a steadier performance across different models, particularly with LongT5, suggesting its robustness in accurate evaluations. Both metrics show weak alignment in assessing readability, reflecting the inherent challenge of one LLM evaluating another's ability to produce easily understandable text. 

\noindent Additionally, LLM-based metrics sometimes fail to align with human judgment, particularly when answers or instructions involve unfamiliar HR terms or sensitive information. Notably, OpenAI models' novel answers exhibit lower human correlation compared to LongT5, which provides answers more similar to the golden response.

\section{Related Work}

Previously, domain-specific chatbots meant for a specific task were designed using conversational AI frameworks like RASA \cite{rasa}. Latest advancements in NLP have shifted focus towards employing and optimizing LLM-based RAG \cite{gao2024retrievalaugmented}. \citet{chen2023benchmarking} experiment with ChatGPT and several other open-source models like Vicuna to benchmark their capabilities in RAG, and \citet{wang2023empower} use a smaller secondary domain-specific model to assist a bigger LLM on a domain-specific question answering task on industrial data. Recent studies have explored various retrieval methods, including dense vector retrieval \cite{dense_passage}, sparse retrieval \cite{bm25,retrieval_techniques}, and hybrid approaches \cite{realm}, to improve the relevance and diversity of retrieved documents. \citet{rag_qa} uses various RAG techniques to ensure that chatbot responses are based on relevant HR policies, leading to accurate and helpful user support.

Given the diverse distribution of the text generated by LLMs, conventional metrics are not suitable for its evaluation \cite{wei2021first,autom_lacking,novikova2017we}. Consequently, a lot of follow-up research has come up in the area of NLG Evaluation \cite{current_status,leveraging}. Specifically focusing on RAG, \citet{es-etal-2024-ragas} released a Framework for the automatic evaluation of generated output using LLM-based metrics with a focus on faithfulness. A similar approach is followed by \citet{saadfalcon2023ares} in their framework ARES which also evaluates the performance of RAG systems over relevance and faithfulness by fine-tuning a lightweight LM judge.

\section{Conclusion}

By optimizing retrieval techniques and benchmarking state-of-the-art LLMs with the help of domain experts, we show how LLM-based applications could benefit from a domain expert as human-in-the-loop within various iterations of the development. Even though our optimizations on the OpenAI-based retriever show minor improvements, the accuracy remains quite low due to the poor quality of the evaluation dataset. Nonetheless, both ChatGPT and GPT-4 show competence when addressing the user query. This hints that the internal reasoning capabilities and domain knowledge of these LLMs are strong enough to overcome the knowledge in the \textit{supposed incorrect article}. This also suggests that, given the nature of the dataset used, the accuracy metric used for the evaluation of the retriever is not a good measure of its performance. We employed and studied a range of evaluation metrics and concluded that in contrast to traditional evaluation approaches such ROUGE \& BERTScore, LLM-based metrics such as Prometheus and G-Eval come very close to human evaluation on average. Nonetheless, our findings reiterate the importance of human judgment, particularly in use cases that require an understanding of a specific domain.

\section*{Acknowledgements}
The work outlined in this paper is part of a research project between the Technical University of Munich and SAP SE under SAP@TUM Collaboration Lab. The authors would like to thank Patrick Heinze, Christopher Pielka, Albert Neumueller, Darwin Wijaya from the SAP IES as well as the Domain Experts from the Human Resource department for their continued support.

\section*{Limitations}
In our experiments, we mostly worked with OpenAI models which are closed-source and hence raise concerns of privacy. Additionally, their large sizes inhibited fine-tuning as they required extensive hardware. Fine-tuning open source and smaller models tailored to HR-specific contexts could further improve response accuracy and relevance. Additionally, since we worked with only one domain expert for the evaluation of the generated answers, the human evaluation might be biased. Because of the data protection concerns with the associated dataset, we cannot make the dataset open source. We employed basic filtering techniques to include user-specific information and context, more advanced approaches could be explored to include this information into the LLM prompt.

\section*{Ethics Statement}
Throughout our experiments, we strictly adhere to the ACL Code of Ethics. The dataset used for our research was anonymized to not include any personal information. We employed in-house domain experts, who receive a full salary for evaluation for generated summaries. They were informed about the task and usability of data in the research. Their annotations were stored in an anonymized fashion, mitigating any privacy concerns. Through our fine-tuning strategies, no additional bias was introduced into the models, other than what might already be part of the dataset. The goal of the research was to optimize an LLM-centric chatbot with the help of a human-in-the-loop. The results and discussions in this paper are meant to further promote research in LLM-based development, bridging the gap between academia and application. 
\bibliography{custom}

\appendix

\section{Dataset}
\subsection{Dataset Collection}
\label{appendix:data-collection}
\textbf{FAQ Dataset: } The internal HR policies of the company consist of Wiki articles, where each article contains a description text followed by some frequently asked questions. The FAQ dataset was constructed by the domain articles by compiling all the FAQ questions from all articles. Each FAQ question is in the form of a triplet where the context is the original Wiki article the question was derived from.
\textbf{UT Dataset: } The user utterance (UT) dataset was compiled using the user utterances collected from the chatbot logs. To reduce the manual labeling effort, a simple text-matching approach was deployed that mapped each user query to one of the questions from the FAQ dataset. The respective answers and context of the matched question were used to create the triplets that form the UT dataset.  
\subsection{Dataset Pre-processing}
\label{app:data-cleaning}
We cleaned the dataset using regular expressions and with the help of LLMs. This involved removing unnecessary formatting like HTML tags, leading or trailing white spaces and newline characters, and removing some wasteful markdown annotations without text. This process thus reduced the number of tokens in each document. Some of the documents were too long to fit into the LLM’s context window, so we excluded them from our analysis.




\subsection{Dataset Challenges}
\label{app:dataset-challenges}

We discovered that our dataset contains multiple articles answering most questions. These articles differ in a few characters, often in an unequal amount of whitespaces, or a few exchanged words, or even entire sections not present in other articles. This situation leads to multiple slightly different versions of the same article present in the dataset, all linked to similar questions. Consequently, the retriever often retrieves very relevant articles that do not exactly match the gold standard article but are a slightly different version. 

To address this, we implemented an evaluation method measuring the Levenshtein distance between the retrieved article and the gold article. If this distance is below a threshold of 100, we consider it a successful retrieval. However, this approach does not match articles with varying sections, as the Levenshtein distance is much higher, and we didn't want to risk matching incorrect articles by increasing the threshold. All of the results in \autoref{tab:evaluation_retriever} are using this evaluation method.

As the DPR is fine-tuned on the dataset, which likely has a strong imbalance in the counts of different article versions, it tends to favor the most common version. This bias contributes to its higher accuracy, as the retriever fetches the correct article more often than not.

\subsection{Dataset Example}
\label{appendix:dataset_example}

\autoref{tab:dataset_sample} shows an example sample from the FAQ dataset representing the training triplet along with all metadata.



\begin{table}[h]
\centering
\resizebox{\columnwidth}{!}{%
\begin{tabular}{p{8cm}l}
\textbf{DATA TRIPLET}\\
\toprule 

\textbf{Question:} How can I apply for half a day of holiday? \\
\textbf{Answer:}  Unfortunately, vacation days in your country can only be taken as full days. \\
\textbf{Context:} \{Relevant Article\} 
\newline
\\
\textbf{META DATA}\\
\toprule
\textbf{User Role:} Employee

\textbf{Name of KBA:} Vacation

\textbf{Company Name:} \{Company Name\}

\textbf{Company Code:} \{Company Code\}

\textbf{Region:} \{Region\}

\textbf{Country Code:} \{Country Code\}

\textbf{FAQ Category:} \{FAQ Category\}

\textbf{Process ID:} \{Process ID\}

\textbf{Service ID:} \{Process ID\}

\end{tabular}
}
\caption{HR Dataset Sample}
\label{tab:dataset_sample}
\end{table}

\section{Human-in-the-Loop}
\label{appendix:human-in-the-loop}
As shown in \autoref{fig:methodology}, the domain experts are involved in various parts of the development cycle explained below:

\noindent\textbf{Dataset Collection:} The domain experts play a big role in the compilation and quality control of the datasets used in this paper

\noindent\textbf{Prompt Optimization: } The domain experts evaluated answers generated by models on various prompt versions. They also provided guidelines the chatbot should follow when addressing the user query which is reflected in the final prompt displayed in \autoref{tab:chatbot_prompt}.

\noindent\textbf{Evaluation:} Domain experts also served as the human annotators for the answers generated by (L)LMs which helped us assess the quality of answers as well as study the effectiveness of automatic evaluation scores.

\section{Prompts Samples}
\label{appendix: prompts}
In this section, we provide the extensive list of prompts used for the OpenAI Models for the Chatbot Pipeline, as well as the prompts used for the LLM-based Metrics.

\subsection{Prompts used for OpenAI models}
\label{chatbot_prmompt}

The optimized prompt used for ChatGPT and GPT-4 during our experiments is shown in \autoref{tab:chatbot_prompt}. 

\begin{table*}[]
\small
\centering
\begin{tabular}{p{15.4cm}l}
\textbf{SYSTEM PROMPT}\\
\toprule 

You are an HR chatbot for SAP SE and you provide truthful and concise answers to employee questions based on provided relevant HR articles.

1. Stay very concise and keep your answer below 150 words.

2. Do not include too much irrelevant information unrelated to the posed question.

3. Keep your response brief and on point.

4. Include URLs from the relevant article if it is important to answer the question.

5. If the answer applies to specific labs/countries/companies, include this information in your response.

6. Refer to the employee directly as "you" and not indirectly as "the employee".

7. If the provided HR article does not include the answer to the question, tell the employee to create an HRdirect ticket.

8. Answer in a polite, personal, user-friendly, and actionable way.

9. Never make up your response! If you do not know the answer to the question, just say so and ask the user to create an HRdirect ticket!
\newline
\\
\textbf{USER PROMPT}\\
\toprule
Question: \{question\}

Relevant Article: \{article\}


\end{tabular}
\caption{Chatbot Prompt for OpenAI Models}
\label{tab:chatbot_prompt}
\end{table*}

\subsection{G-Eval Evaluation Metric Prompt}
The evaluation prompt used for the Readability Criteria is shown in \autoref{tab:g-eval_prompt}. The prompts for other criteria (Truthfulness, Usability, Relevance) follow similar instructions as the one shown for the Readability prompt. 

\begin{table*}[]
\small
\centering
\begin{tabular}{p{15.4cm}l}
\textbf{SYSTEM PROMPT}\\
\toprule 
You will be given a generated answer for a given question. Your task is to act as an evaluator and compare the generated answer with a reference answer on one metric. The reference answer is the fact-based benchmark and shall be assumed as the perfect answer for your evaluation. Please make sure you read and understand these instructions very carefully. Please keep this document open while reviewing, and refer to it as needed.

Evaluation Criteria: \{criteria\}

Evaluation Steps: \{steps\}
\newline
\\
\textbf{USER PROMPT}\\
\toprule
Example: \{example\}

Question: \{question\}

Generated Answer: \{generated\_answer\}

Reference Answer: \{reference\_answer\}

Evaluation Form: Please provide your output in two parts separate as a Python dictionary with keys rating and explanation. 
First the rating in an integer followed by the explanation of the rating.

\{metric\_name\}
\newline
\\
\textbf{METRIC SCORE CRITERIA}\\
\toprule
\{The degree to which the generated answer matches the reference answer based on the metric description.\}

Readability(1-5) - Please rate the readability of each chatbot response. This criterion assesses how easily the response can be understood. A response with high readability should be clear, concise, and straightforward, making it easy for the reader to comprehend the information presented. Complex sentences, jargon, or convoluted explanations should result in a lower readability score.
\newline
\\
\textbf{METRIC SCORE STEPS}\\
\toprule
\{Readability Score Steps\}

1. Read the chatbot response carefully.

2. Assess how easily the response can be understood. Consider the clarity and conciseness of the response.

3. Consider the complexity of the sentences, the use of jargon, and how straightforward the explanation is.

4. Assign a readability score from 1 to 5 based on these criteria, where 1 is the lowest (hard to understand) and 5 is the highest (very easy to understand).

\end{tabular}
\caption{G-Eval Prompt Example for Readability Criteria}
\label{tab:g-eval_prompt}
\end{table*}

\newblock \subsection{Prometheus Evaluation Metric Prompt}

The prompt for the Prometheus Evaluation Metric outlined in \autoref{tab:prometheus_prompt} was based on the official paper’s guidelines \cite{prometheus} for Feedback Collection. This specific prompt illustrates the Readability Criteria and was similarly adapted for other criteria such as Truthfulness, Relevance, and Usability. In general, both LLM-based metrics follow similar evaluation criteria in the prompts.

\begin{table*}[]
\small
\centering
\begin{tabular}{p{15.4cm}l}
\textbf{SYSTEM PROMPT}\\
\toprule 
\textbf{Task Description:} 
An instruction (might include an input inside it), a response to evaluate, a reference answer that gets a score of 5, and a score rubric representing an evaluation criterion is given. \\
2. After writing a feedback, write a score that is an integer between 1 and 5. You should refer to the score rubric. \\
3. The output format should look as follows: Feedback: [write a feedback for criteria] [RESULT] [an integer number between 1 and 5]. \\
4. Please do not generate any other opening, closing, and explanations. \\
\midrule
\textbf{Question to Evaluate:} \{instruction\} \\
\textbf{Response to Evaluate:} \{response\} \\
\textbf{Reference Answer (Score 5):} \{reference answer\} \\
\textbf{Score Rubrics:} \{criteria description\} \\
Score 1: \{Very Low correlation with the criteria description\} \\
Score 2: \{Low correlation with the criteria description\} \\
Score 3: \{Acceptable correlation with the criteria description\} \\
Score 4: \{Good correlation with the criteria description\} \\
Score 5: \{Excellent correlation with the criteria description\} \\
\textbf{\{criteria description\}}: Readability(1-5) - Please rate the readability of each chatbot response. \
This criterion assesses how easily the response can be understood. A response with high readability should be clear, concise, and straightforward. Complex sentences, jargon, or convoluted explanations should result in a lower readability score.

\end{tabular}
\caption{Prometheus Prompt Example for Readability Criteria}
\label{tab:prometheus_prompt}
\end{table*}

\newcommand{\redact}[1]{\textcolor{black}{\textbf{[REDACTED]}}}

\section{Technical Details}
\subsection{Retriever}
\label{app:dpr-training}
It is worth noting that we embed the whole article and do not perform chunking. As shown in \autoref{fig:n_tokens_articles}, these articles are quite long. To cater to the limited context window of the models, we opt for the top-1 article to be passed as context. This also makes sense for our use case as the dataset is designed such that the answer to any given HR question usually exists in only one article.

\subsection{Dense Passage Retriver Training}
Dense Passage Retriever (DPR) \cite{karpukhin2020dense} powered by Haystack\footnote{\url{https://haystack.deepset.ai/}} uses the \textit{bert-base-uncased} embedding model by \textit{google-bert}, openly available on HuggingFace. DPR training aims to generate a model that creates embeddings where the question embedding closely aligns with the relevant context embedding. During retrieval, the user query is processed through the previously trained retriever, producing a query vector in the same embedding space as the articles. This query vector is then compared to all article vectors within the vector store using cosine similarity. The top-k articles belonging to the embeddings with the highest cosine similarities are returned.
\subsection{LongT5 Fine-tuning}
During fine-tuning of the LongT5 models, the training process was configured with a learning rate of 1e-4 and a batch size of 8, spanning 5 epochs.

\section{Results and Evaluation}
Throughout our research, we encountered several challenges that warrant attention. The variability in retrieved articles due to slight differences in content or formatting posed complexities in evaluating retrieval accuracy and ensuring consistency in response generation. Addressing this challenge may require further refinement of the retrieval mechanism or additional preprocessing steps to standardize the retrieved content.

\subsection{Retriever}
\label{sec:topk-evaluation_retriever}
The accuracy of both DPR on the top-1, top-2, top-3, and top-5 articles on both retrievers is shown in \autoref{tab:topk-evaluation_retriever}. As expected, the accuracy of the retriever module increases as the value of k is increased. However, we are limited to including only top-1 articles because the articles are quite long and more samples may not fit in the model's context window. The BERT-based DPR model still significantly outperforms all new methods with a top-1 accuracy of 22.24\% and a top-5 accuracy exceeding 40\%. The new retriever, in comparison, only reaches a top-1 accuracy of 11.12\% and a top-5 accuracy of 18.53\% on the same dataset. These results in general are quite underwhelming and mainly attributed to the dataset challenges described in Appendix \ref{app:dataset-challenges}.
\begin{table}[H]
\centering
\resizebox{\columnwidth}{!}{%
\begin{tabular}{c|cccc}
\textbf{DPR} & \textbf{top-1}  & \textbf{top-2} & \textbf{top-3} & \textbf{top-5} \\
\hline
BERT-based & 22.24\% & 30.03\% & 35.08\% & \textbf{40.06\%} \\
\hline
OpenAI-based & 11.12\% & 15.06\% & 16.82\% & \textbf{18.53\%} \\

\hline
\end{tabular}
}
\caption{Retriever Accuracy on the HR test dataset for various values of k on the HR Dataset. The OpenAI-based DPR uses the \textit{Basic} method.}
\label{tab:topk-evaluation_retriever}
\end{table}

\subsection{Correlation between Automatic Evaluation and Domain Expert Evaluation} 
\autoref{tab:metrics-correlation} shows the individual across for correlation of each evaluation metric with human evaluation across LongT5, ChatGPT, and GPT-4. The low correlation coefficients are a consequence of the Spearman and Kendall methods, which analyze the linear and ordinal relationships between variables by comparing each set of scores. When these methods detect divergent scores between two evaluations, it leads to a reduced correlation coefficient, indicating a disproportion that is not apparent when considering the average scores alone.

\begin{table*}[!ht]
\centering
\begin{tabular}{lS[table-format=1.4]S[table-format=1.4]S[table-format=1.4]S[table-format=1.4]S[table-format=1.4]S[table-format=1.4]}
\toprule
{Criteria} & \multicolumn{2}{c}{LongT5} & \multicolumn{2}{c}{ChatGPT} & \multicolumn{2}{c}{GPT-4} \\
\cmidrule(lr){2-3} \cmidrule(lr){4-5} \cmidrule(lr){6-7}
& {Spearman \( \rho \)} & {Kendall \( \tau \)} & {Spearman \( \rho \)} & {Kendall \( \tau \)} & {Spearman \( \rho \)} & {Kendall \( \tau \)} \\
\midrule
\addlinespace 
BLEU           & \textbf{0.459}  & 0.337  & 0.345 & 0.263 & 0.146 & 0.116 \\
ROUGE-1         & \textbf{0.435}  & 0.321  & 0.364 & 0.284 & 0.113 & 0.091 \\
ROUGE-2         & \textbf{0.462}  & 0.341  & 0.332 & 0.258 & 0.056 & 0.044 \\
ROUGE-L         & \textbf{0.433}  & 0.324  & 0.353 & 0.274 & 0.093 & 0.075 \\
BERTScore\_P    & \textbf{0.457}  & 0.347  & 0.304 & 0.234 & 0.156 & 0.122 \\
BERTScore\_R    & \textbf{0.466}  & 0.305  & 0.085 & 0.064 & -0.022 & -0.018 \\
BERTScore\_F1   & \textbf{0.455}  & 0.332  & 0.246 & 0.192 & 0.097 & 0.077 \\

\textbf{G-Eval} \\
Usability       & 0.675 & 0.584 & 0.217 & 0.198 & 0.346 & 0.327 \\
Relevance       & 0.569 & 0.499 & 0.339 & 0.304 & 0.325 & 0.306 \\
Readability     & 0.208 & 0.181 & 0.395 & 0.373 & 0.139 & 0.137 \\
Truthfulness    & \textbf{0.726} & 0.651 & 0.694 & 0.667 & 0.452 & 0.432 \\
\addlinespace 
\textbf{Prometheus} \\
Usability       & \textbf{0.723} & 0.675 & 0.386 & 0.351 & 0.516 & 0.495 \\
Relevance       & 0.467 & 0.439 & 0.419 & 0.371 & 0.382 & 0.357 \\
Readability     & 0.493 & 0.468 & 0.378 & 0.358 & 0.225 & 0.213 \\
Truthfulness    & 0.541 & 0.521 & 0.439 & 0.402 & 0.454 & 0.427 \\
\bottomrule
\end{tabular}

\centering
\caption{Correlations between Automated Metrics and Human Evaluation across Models}
\label{tab:metrics-correlation}

\end{table*}



\end{document}